%% file: SCIS-2021-1065.tex
\documentclass{SCIS2025}

\usepackage[english]{babel}
\usepackage{diagbox}
\usepackage{multirow}%
\usepackage{xcolor}%
\usepackage{textcomp}%
\usepackage{manyfoot}%
\usepackage{booktabs}%
\usepackage[colorlinks,linkcolor=black]{hyperref}
\newtheoremstyle{exampstyle}
{0.0em} 
{0.0em} 
{} 
{1em} 
{\bfseries} 
{.} 
{1em} 
{} 
\usepackage{balance}
\theoremstyle{exampstyle}

\usepackage{CJKutf8}


\begin{document}
	\ArticleType{RESEARCH PAPER}
	\Year{2025}
	\Month{November}
	\Vol{}
	\No{}
	\DOI{}
	\ArtNo{}
	\ReceiveDate{}
	\ReviseDate{}
	\AcceptDate{}
	\OnlineDate{}
	
	\title{DEAR: Dataset for Evaluating the Aesthetics of Rendering}{DEAR: Dataset for Evaluating the Aesthetics of Rendering}
	
	\author[1, 2, 3]{Vsevolod Plohotnuk}{{plokhotnyuk.v@miriai.org}}
    \author[1, 2]{Artyom Panshin}{}
    \author[4]{Nikola Bani\'{c}}{}
    \author[5]{Simone Bianco}{}
    \author[6]{\\Michael Freeman}{}
    \author[1, 2, 3]{Egor Ershov}{}
	
	
	
    
    \address[1]{Color Reproduction and Synthesis Institute, Moscow {\rm 119991}, Russia}
    \address[2]{Moscow Independent Research Institute of Artificial Intelligence, Moscow {\rm 125635}, Russia}
    \address[3]{Computational color photography Laboratory, Moscow {\rm 123112}, Russia}
    \address[4]{Gideon Brothers, Zagreb {\rm 10000}, Croatia}
    \address[5]{University of Milano-Bicocca, Milan {\rm 20126}, Italy}
    \address[6]{Michael Freeman Photography}
	
    
\abstract{Traditional Image Quality Assessment~(IQA) focuses on quantifying technical degradations such as noise, blur, or compression artifacts, using both full-reference and no-reference objective metrics. 
However, evaluation of rendering aesthetics, a growing domain relevant to photographic editing, content creation, and AI-generated imagery, remains underexplored due to the lack of datasets that reflect the inherently subjective nature of style preference. 
In this work, a novel benchmark dataset designed to model human aesthetic judgments of image rendering styles is introduced: the Dataset for Evaluating the Aesthetics of Rendering (DEAR).
Built upon the MIT-Adobe FiveK dataset, DEAR incorporates pairwise human preference scores collected via large-scale crowdsourcing, with each image pair evaluated by $25$ distinct human evaluators with a total of $13,648$ of them participating overall. 
These annotations capture nuanced, context-sensitive aesthetic preferences, enabling the development and evaluation of models that go beyond traditional distortion-based IQA, focusing on a new task: Evaluation of Aesthetics of Rendering (EAR). 
The data collection pipeline is described, human voting patterns are analyzed, and multiple use cases are outlined, including style preference prediction, aesthetic benchmarking, and personalized aesthetic modeling. 
To the best of the authors' knowledge, DEAR is the first dataset to systematically address image aesthetics of rendering assessment grounded in subjective human preferences.
A subset of 100 images with markup for them is published on HuggingFace.\footnote{\href{https://huggingface.co/datasets/vsevolodpl/DEAR}{https://huggingface.co/datasets/vsevolodpl/DEAR}}
}


\maketitle

\Acknowledgements{The work was supported by the Ministry of Economic Development of the Russian Federation (agreement No. 139-15-2025-013, dated June 20, 2025, IGK 000000C313925P4B0002).}

\begin{figure}[ht]
    \centering
    \includegraphics[width=0.75\linewidth]{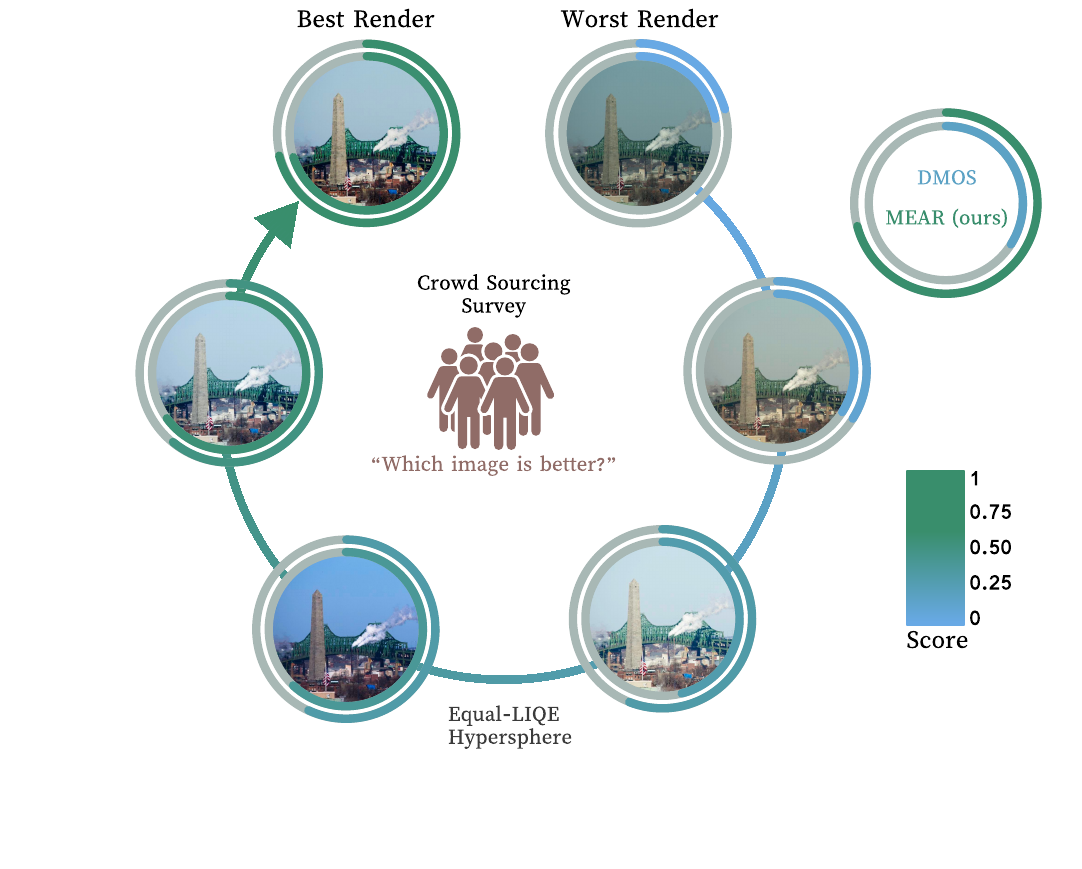}
    \caption{Limitations of existing IQA methods: while LIQE \cite{Zhang2023} assigns nearly identical quality scores with standard deviations from the mean value 3.5\% across evaluations, subjective evaluation reveals clear differences.
    Our work addresses this gap by introducing a new benchmark  Dataset for Evaluating of Aesthetics of Rendering (DEAR).}
    \label{fig:abstract}
\end{figure}
\input{intro}
\input{related_work}
\input{dataset}
\input{Analysis_of_preferences}

\input{quality_prediction}

\input{discussion}

\input{concl}
\bibliographystyle{splncs04}
\bibliography{sn-bibliography}

\input{supplementary}
\end{document}

%% file: intro.tex
\section{Introduction}
\label{sec:introduction}
Image Quality Assessment refers to the task of automatically evaluating the quality of images in a way that aligns with human perception. 
Accurate IQA is critical for the development and optimization of computer vision and image processing algorithms, as it provides an objective signal that guides models toward producing results that are visually pleasing or perceptually faithful~\cite{zhai2020perceptual}.
Whether in image compression, enhancement, restoration, or generation, IQA metrics are expected to help ensure that algorithmic outputs meet human expectations of quality, as well as to measure the performance of the algorithms themselves.

Traditional IQA research has focused primarily on assessing the degradation of perceived image quality introduced by the presence of distortions such as noise, blur, or compression artifacts.
Over time, however, the field recognized that perceived image quality extends beyond simple degradations: viewers judge an image not only by the absence of artifacts, but also by how convincingly it conveys realism and visual appeal. 
This shift in perspective motivated the creation of modern subjective IQA datasets, such as UHD-IQA~\cite{Hosu2024}, which capture human judgments on more holistic attributes, including composition, resolution, and color harmony, without requiring a pristine reference.
However, even in subjective IQA datasets~\cite{Hosu2024}, participants are typically asked to rate the overall quality of individual images, not to compare different stylistic renditions of the same scene.
As a result, such datasets capture judgments about technical fidelity or perceptual realism rather than aesthetics of rendering, and are not suited for modeling user preferences between alternative rendering interpretations, which could be captured much more easily if datasets such as UHD-IQA contained pairwise evaluation results.

Evaluation of aesthetics of rendering introduces a qualitatively different challenge.
Aesthetics of rendering is not characterized by degradations, but by aesthetic attributes such as color usage, texture, and artistic expression.
Importantly, while image degradations are known to be of lesser quality than the original image, multiple renderings can be equally valid and pleasing, making rendering preference inherently subjective and context-dependent.
Unlike quality degradations, which tend to have a broadly agreed-upon negative impact, rendering variation often reflects personal, cultural, or situational preferences.
Finally, because there is no objectively correct answer for the most pleasant rendering, it is impossible to guarantee whether users select their responses attentively, making the annotation process vulnerable to careless behavior~\cite{ershov2024reliability}.

Despite growing interest in modeling visual aesthetics and style, the field lacks robust datasets that reflect the subjective nature of style perception.
Existing IQA datasets are not suitable for learning or evaluating models that aim to predict human style judgments, as they are designed for scenarios with clear perceptual rankings, see Figure~\ref{fig:abstract}.
This absence of EAR data hinders progress in developing algorithms that can interpret and predict nuanced aesthetic preferences.

To address this gap, we introduce the Dataset for Evaluating the Aesthetics of Rendering (DEAR), a new benchmark specifically designed to allow learning subjective judgments of image rendering.
DEAR is constructed by augmenting the MIT-Adobe FiveK dataset~\cite{fivek} with pairwise human preference scores obtained through crowdsourced evaluation platform.
Each pair of image renderings was evaluated by at least $25$ evaluators, with a total of $13,648$ unique evaluators participating overall.
These annotations enable the development of EAR models that go beyond degradation-focused assessments and instead capture the more nuanced, human-centered perception of rendering.

The dataset includes both the complex, photographer-specific rendering variations originally present in the MIT-Adobe FiveK dataset, as well as simpler rendering modifications such as subtle adjustments in contrast.
This diversity allows DEAR to reflect a broader and more realistic range of aesthetic choices.
To the best of our knowledge, this is the first dataset of its kind, making it possible to conduct experiments on subjective rendering perception that were previously impractical or infeasible.

The paper is structured as follows: Section~\ref{sec:related} gives a preview of the related work, in Section~\ref{sec:proposed} the new Dataset
for Evaluating the Aesthetics of Rendering is described, in Section~\ref{sec:preferences} the analysis of users' preferences is performed, Section~\ref{sec:prediction} describes some new models for EAR and their quality, in Section~\ref{sec:discussion} the results are presented and discussed, and Section~\ref{sec:conclusions} concludes the paper.

%% file: related_work.tex
\section{Related work}
\label{sec:related}

The section on related work is organized into two key subsections, aligning with the paper's core contributions.
First, existing research on the development of standardized benchmark databases for IQA is comprehensively reviewed.
Second, the state-of-the-art methods for no-reference (NR) IQA are critically analyzed.
This dual-pronged approach enables a systematic examination of both (1) the foundational datasets essential for reliable evaluation and (2) the most advanced techniques driving progress in NR-IQA.

\subsection{Benchmark IQA datasets}

In recent years, numerous IQA datasets have been introduced. 
Benchmarks such as LIVE~\cite{sheikh2005live}, VCL@FER~\cite{Zari2012}, NITS-IQA~\cite{Ruikar2023}, and CSIQ~\cite{Chandler2010} comprised limited sets of images with diverse distortion types, annotated using controlled laboratory studies. 
Subsequent efforts such as TID2013~\cite{Ponomarenko2015} and MDID~\cite{Sun2017} expanded dataset sizes but remained insufficient for training data-hungry deep learning models due to their inherent scale limitations. 
A key bottleneck is the labor-intensive and costly subjective scoring process, which hinders the creation of large-scale datasets. 
As a result, only a few sizable datasets such as PDAP-HDDS~\cite{Liu2018} exist.
Moreover, synthetically marked-up datasets also exist, e.g., KADIS-700k~\cite{Lin2020} with markups obtained by a model trained on KADID-10k~\cite{Lin2020}, but unfortunately their utility is compromised for certain evaluation tasks.

Recent studies~\cite{7460955,hosu2018expertise} demonstrate that, with proper experimental design, crowdsourced annotations can achieve a reliability comparable to laboratory settings.
Moreover, in prior work~\cite{ershov2024reliability}, the reliability and stability of using crowdsourcing for stylistic and rendering evaluation was investigated, demonstrating its effectiveness as a viable data collection method.
Crowdsourcing is becoming increasingly important and relying on it has enabled the creation of large-scale, diverse IQA datasets such as KonIQ-10k~\cite{Hosu2019} and UHD-IQA~\cite{Hosu2024}, which address the scalability limitations of traditional approaches.
Unlike earlier benchmarks constrained by small sample sizes or missing subjective labels, these datasets leverage crowdsourcing to balance scale and annotation quality, making them more suitable for modern data-driven IQA methods.

The aforementioned works primarily address objective IQA, focusing on technical distortions such as noise or compression artifacts.
In contrast, a separate line of research targets image aesthetics evaluation, which evaluates subjective preferences and perceptual appeal.
Prominent datasets in this domain include AVA~\cite{murray2012ava}, AVA-PD~\cite{kairanbay2018towards}, and EVA~\cite{kang2020eva}. 
Unlike technical IQA benchmarks, these datasets evaluate aesthetics by comparing entirely different scenes, where users select the most visually pleasing image within a broad theme, e.g., ``flying bird''.
Crucially, however, this evaluation captures holistic judgments incorporating content, composition, lighting, and stylistic choices, rather than isolated technical quality.
This makes it difficult to assess the contribution of stylization or rendering to overall image quality.

Recently, a growing research trend has emerged to evaluate the quality of AI-generated images, addressing unique challenges such as perceptual realism, semantic consistency, and technical artifacts.
Datasets like HPDv2~\cite{wu2023human}, AGIQA-3k~\cite{Li2023}, AIGIQA-20k~\cite{li2024aigiqa}, and PKU-I2IQA~\cite{yuan2023pku} have been developed to assess these aspects.
These datasets typically employ a structured methodology: multiple images are generated for each prompt, and then their quality, authenticity, and prompt alignment are rigorously validated using mean opinion scores.
This approach provides a standardized framework to assess the rapidly evolving capabilities of generative AI models.

Several tone-mapped image quality datasets have as well been introduced~\cite{ak2022rv,jiang2025dataset}, in which multiple tone mapping operators~(TMOs) are applied to high dynamic range~(HDR) images to produce low dynamic range~(LDR) outputs of varying perceptual quality (we use the term HDR in the traditional sense of a high-range image destined for reduction to SDR --- Standard Dynamic Range --- rather than the newer display technology that presents actual HDR to the viewer). While TMOs do not introduce degradations in the traditional sense, making these datasets somewhat aligned with the subjective nature of our work, they differ fundamentally in scope and application. Specifically, the input and output images in such datasets belong to different dynamic ranges, making the transformations more technical than stylistic. Moreover, these datasets do not address or address minimally the broader and more common use cases involving stylistic and rendering variation within the same dynamic range, such as color grading, contrast adjustments, or photographic style emulation, which are the core of EAR and the focus of this work.

The UHD-IQA Benchmark Database~\cite{Hosu2024} contains 6,073 ultra-high-definition~(4K) images collected from Pixabay and curated to exclude synthetic or heavily retouched content. Image quality was assessed in a single-stimulus protocol, where 10 expert raters composed of photographers and graphic artists judged each image twice on a continuous scale from \emph{bad} (1\%) to \emph{excellent} (100\%), yielding around 20 ratings per image. The resulting mean opinion scores provide ground-truth annotations, with high inter-round consistency measured by the Spearman Rank Order Correlation Coefficient (SRCC), reported at approximately $0.93$. Along with perceptual quality ratings, the dataset includes metadata such as image dimensions, popularity indicators, and machine-generated tags, making it one of the largest and most comprehensive publicly available benchmarks for no-reference image quality assessment in the UHD domain. However, the dataset contains only a single rendering per image, limiting its usefulness for evaluating different renderings of the same scene; moreover, the absence of pairwise comparisons further limits its ability to capture fine-grained perceptual differences.

Overall, none of the aforementioned works explicitly addresses EAR. 
While existing approaches focus on quantifying technical distortions --- such as noise, blur, or compression artifacts --- that degrade generic image quality, they largely overlook the perceptual impact of artistic style, see Table~\ref{tab:ds_summary}.
The goal of this paper is exactly to fill this niche.

\begin{table*}[t!]
\centering
\caption{Dataset summary for most relevant IQA datasets compared to our EAR dataset. In this table in the method column ACR stands for 5-point Absolute Category Rating, SS for Single Stimulus, and PC for pairwise comparison.}
\begin{tabularx}{\linewidth}{>{\hsize=1.4\hsize\arraybackslash}X
    >{\hsize=0.8\hsize\centering\arraybackslash}X 
    >{\hsize=0.8\hsize\centering\arraybackslash}X 
    >{\hsize= 1.5 \hsize\centering\arraybackslash}X 
    >{\hsize=0.8\hsize\centering\arraybackslash}X 
    >{\hsize=0.7\hsize\centering\arraybackslash}X}
\toprule
Dataset     & \# Scenes & \begin{tabular}[c]{@{}c@{}}\# Edited\\ Images\end{tabular}  & IQA Task  & \# Voters & Method \\
\midrule
LIVE~\cite{sheikh2005live}          & 29        & 779               & Distortion   & 29        & SS \\
CSIQ~\cite{Chandler2010}            & 30        & 866               & Distortion   & 25        & N/A \\
VCL@FER~\cite{Zari2012}             & 23        & 552               & Distortion   & 118       & SS \\
TID2013~\cite{Ponomarenko2015}      & 25        & 3,000              & Distortion   & 971       & PC\\
PDAP-HDDS~\cite{Liu2018}            & 250       & 12,000             & Distortion   & 38        & ACR \\ 
KonIQ-10k~\cite{Hosu2019}           & 10,073     & 0                 & Distortion+Content            & 1,459      & ACR      \\
KADID-10k~\cite{Lin2020}            & 81        & 10,125             & Distortion   & 2,209      & ACR      \\
UHD-IQA~\cite{Hosu2024}             & 6,073      & 6,073              & Content   & 10      & SS      \\ \midrule
DEAR                                & 5,000      & 30,000             & EAR       & 13,648     & PC\\
\bottomrule
\end{tabularx}
\label{tab:ds_summary}
\end{table*}

\subsection{Benchmark NR-IQA methods}

For IQA tasks, numerous methodological approaches exist.
The field of IQA is dominated by two primary approaches: full-reference and no-reference methods.
Full-reference techniques require a complete, high-fidelity reference image for comparison to quantify perceptual degradation.
Conversely, no-reference methods operate without any reference image, directly estimating the quality score based solely on the features of the test image.
In our work, since we propose a dataset without reference images (i.e., with no ``golden standard'' for comparison exists), traditional full-reference metrics like SSIM~\cite{wang2004image}, PSNR, or LPIPS~\cite{zhang2018unreasonable} cannot be directly applied.
Consequently, our analysis focuses exclusively on NR-IQA methods.

NR-IQA approaches can be broadly categorized into two paradigms:
\begin{itemize}
    \item \textbf{Statistical} methods, that employ handcrafted heuristics to detect specific artifacts (e.g., noise, blur, compression artifacts) based on measurable image statistics.
    These methods typically rely on domain knowledge about common distortion types and their observable effects on image properties.
    \item \textbf{Deep learning} (DL) methods, that utilize data-driven feature extraction to automatically learn complex quality-relevant patterns.
    Unlike statistical approaches, DL models can adapt to image content and capture more sophisticated quality attributes through hierarchical feature learning.
    This enables them to handle diverse distortion types and content variations more effectively.
\end{itemize}

One of the most widely used statistical approaches for NR-IQA is BRISQUE~\cite{Mittal2012}.
This method extracts a set of handcrafted features based on natural scene statistics and employs a Support Vector Regressor (SVR)~\cite{scholkopf2000new} to predict the final quality score.
More recently, PIQUE~\cite{pique} has adopted a similar paradigm, leveraging perceptually relevant features in conjunction with a statistical learning framework to estimate image quality.

With the emergence of large-scale IQA datasets, DL-based methods have gained significant traction.
Early works such as DeepBIQ~\cite{bianco2018use},  MUSIQ~\cite{ke2021musiq}, PaQ-2-PiQ~\cite{ying2020patches}, CONTRIQUE~\cite{madhusudana2022image}, Re-IQA~\cite{saha2023re}, and ARNIQA~\cite{arniqa} explore various deep learning paradigms, including patch-based learning, contrastive representation learning, and transformer-based architectures, to improve the accuracy and generalizability of IQA models.

Furthermore, a growing number of recent approaches incorporates vision-language models for image quality assessment.
For instance, CLIP-IQA~\cite{Wang2023} estimates image quality by computing the cosine similarity between visual embeddings and text prompts such as ``This photo is good" and ``This photo is bad."
Building upon this idea, LIQE~\cite{Zhang2023} improves the framework by fine-tuning both visual and textual encoders and introducing a multi-task optimization strategy, which enhances robustness, particularly due to its use of large-scale training data.
Subsequently, QualiCLIP~\cite{Agnolucci2024} was designed to be more sensitive to varying levels of image distortion.
Most recently, LAR-IQA~\cite{lariqa} demonstrated that incorporating KANs~\cite{liu2024kan} can further boost model performance.

%% file: dataset.tex
\section{Proposed DEAR}
\label{sec:proposed}

The MIT-Adobe FiveK dataset~\cite{fivek} is a widely-used, high-resolution image dataset designed for research in computational photography and image enhancement. 
It comprises 5,000 RAW photographs captured by DSLR cameras under diverse lighting and scene conditions. 
Each image has been independently rendered by five professional photo renderers, labeled Expert A through E, yielding five stylistically distinct reference renders per image. 
The dataset includes both the original RAW files and their corresponding JPEG renderings, making it suitable for tasks such as supervised learning of image-to-image translation, automatic photo retouching, tone and exposure correction, white balance adjustment, style-specific enhancement, etc. 
Its diversity in photographic content, expert renderers, and high-resolution format with up to $5616\times 3744$ pixels makes it a valuable resource for training and evaluating deep learning models in both paired and unpaired settings.

\begin{figure}
    \centering
    \includegraphics[width=0.75\linewidth]{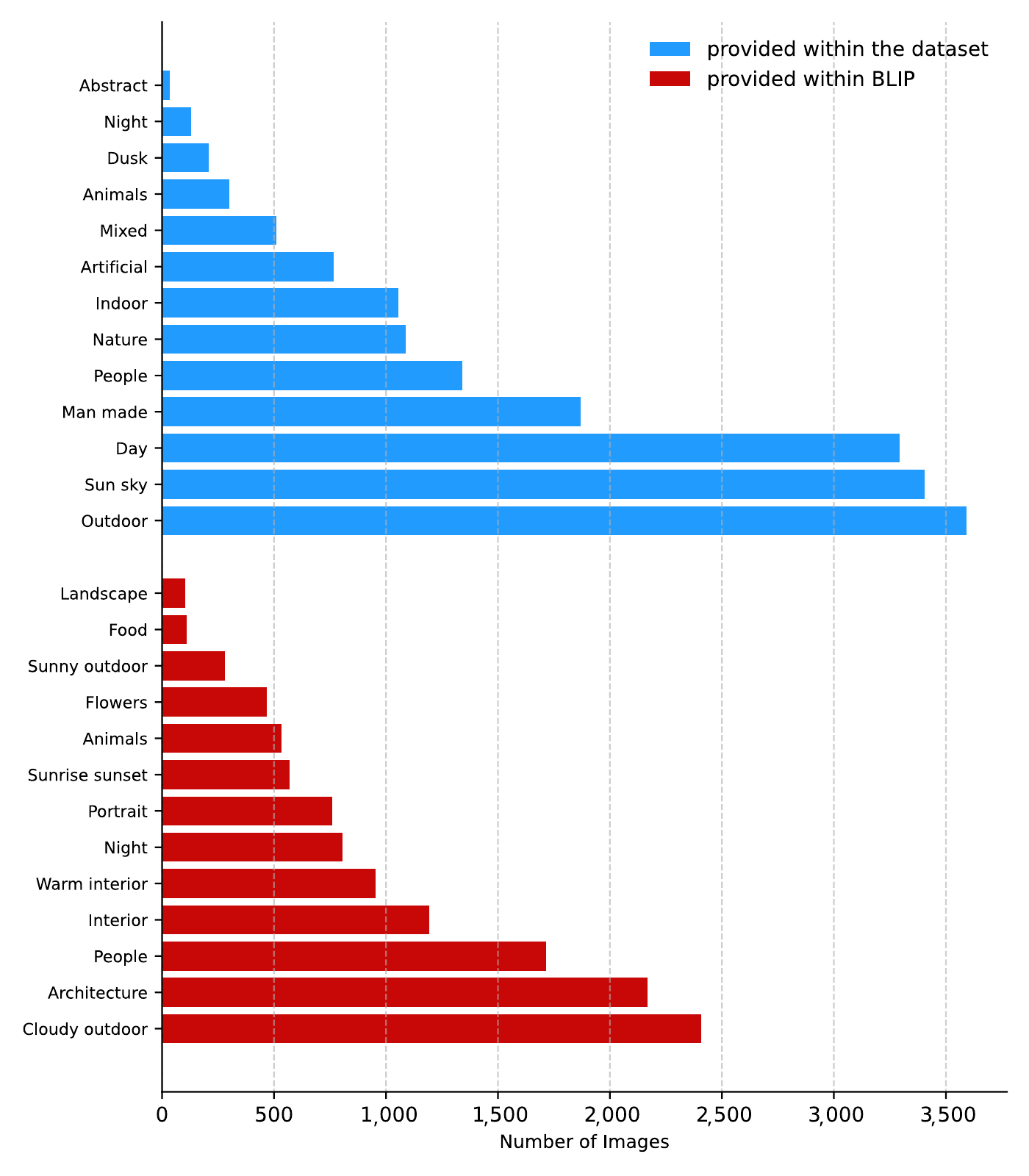}
    \caption{The distribution of categories for the selected MIT-Adobe FiveK dataset images based on the information provided within the dataset and BLIP; one image can have multiple categories.}
    \label{fig:categories_dataset}
\end{figure}

The distribution of labels for these images that is available within the MIT-Adobe FiveK dataset is shown in Figure~\ref{fig:categories_dataset}, together with another distribution obtained by means of querying the BLIP~\cite{li2022blip}. 
In order to validate the overall quality of BLIP results we randomly selected 100 images from the dataset and marked them up.
The average accuracy for all categories equals $0.96$, while F1-score equals $0.86$.
More detailed analysis of category markup can be seen in the Supplementary material.
The variety of the dataset is sufficient to draw general conclusions.

\begin{figure}[htb]
    \centering
    
	\includegraphics[width=0.7\linewidth]{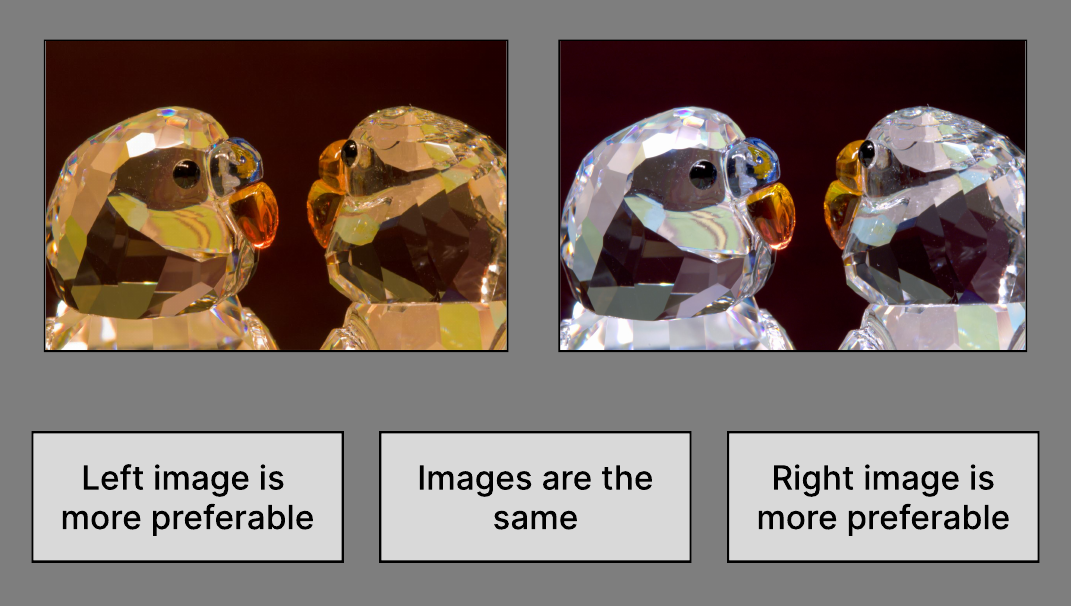}
	
    \caption{An example of a pair of rendered styles for the same scene as shown in Yandex.Tasks.}
	\label{fig:pair}
    
\end{figure}

Each of the MIT-Adobe FiveK images is rendered in six different styles: five photographers, i.e. renderers and one neutral style from Adobe Photoshop {\it{auto}} settings, which yields $\binom{6}{2} = 15$ possible unique style pair combinations per image.
To perform subjective quality assessment, these pairs were uploaded to the crowdsourcing platform Yandex.Tasks.
The Yandex.Tasks interface displayed six image pairs per page.
For each pair, the participants, i.e., evaluators were asked: \textit{``Which image is preferable?''}.
Images were presented against a 50\% gray background, following widely used recommendations~\cite{mantiuk2012comparison} and standard to minimize bias in color perception by using a neutral background.
To further ensure unbiased assessment, no additional controls or information were placed between the image pairs. Each image pair was also spaced adequately to simplify the visual comparison.
An example of such an image pair can be seen in Figure~\ref{fig:pair}.

The evaluators could select one of the following options for each image pair: ``The left image is preferable'', ``The right image is preferable'' and ``Both images are equally preferable''.
To identify unreliable evaluators, each page included an additional pair of nearly identical images as a control question. Essentially, the whole voting process was similar to the one in~\cite{ershov2024reliability}.

\begin{figure}[!ht]
    \centering
    \includegraphics[width=\linewidth]{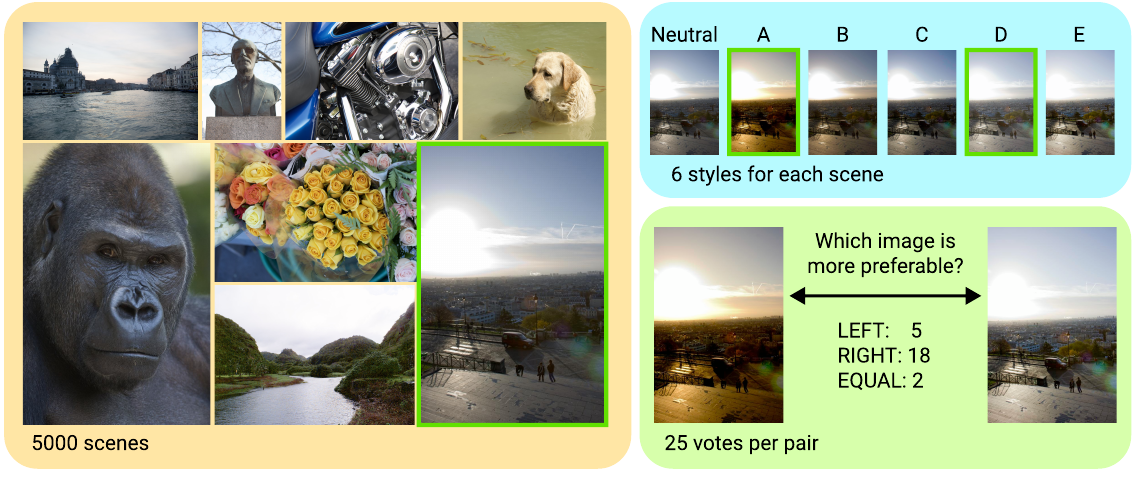}
    \caption{Illustration of the proposed DEAR dataset. For each scene in MIT-Adobe FiveK there are 6 different styles: 5 styles from renderers and 1 for Adobe neutral style. For each pair in scene there is a total of 25 votes subdivided between left image, right image, and equal.}
    \label{fig:alldata}
\end{figure}

The dataset consists of 5,000 unique scenes captured by professional photographers.
For each scene, five rendered versions were produced by photography students, along with one automatically generated version using Adobe Photoshop’s auto-mode.
For EAR, each pair of scene versions was assessed by 25 participants, who provided one of three possible responses: a preference for one image, a preference for the other, or abstention.
An illustration of the dataset is shown of Figure~\ref{fig:alldata}. 

\begin{figure}
    \centering
    \includegraphics[width=\linewidth]{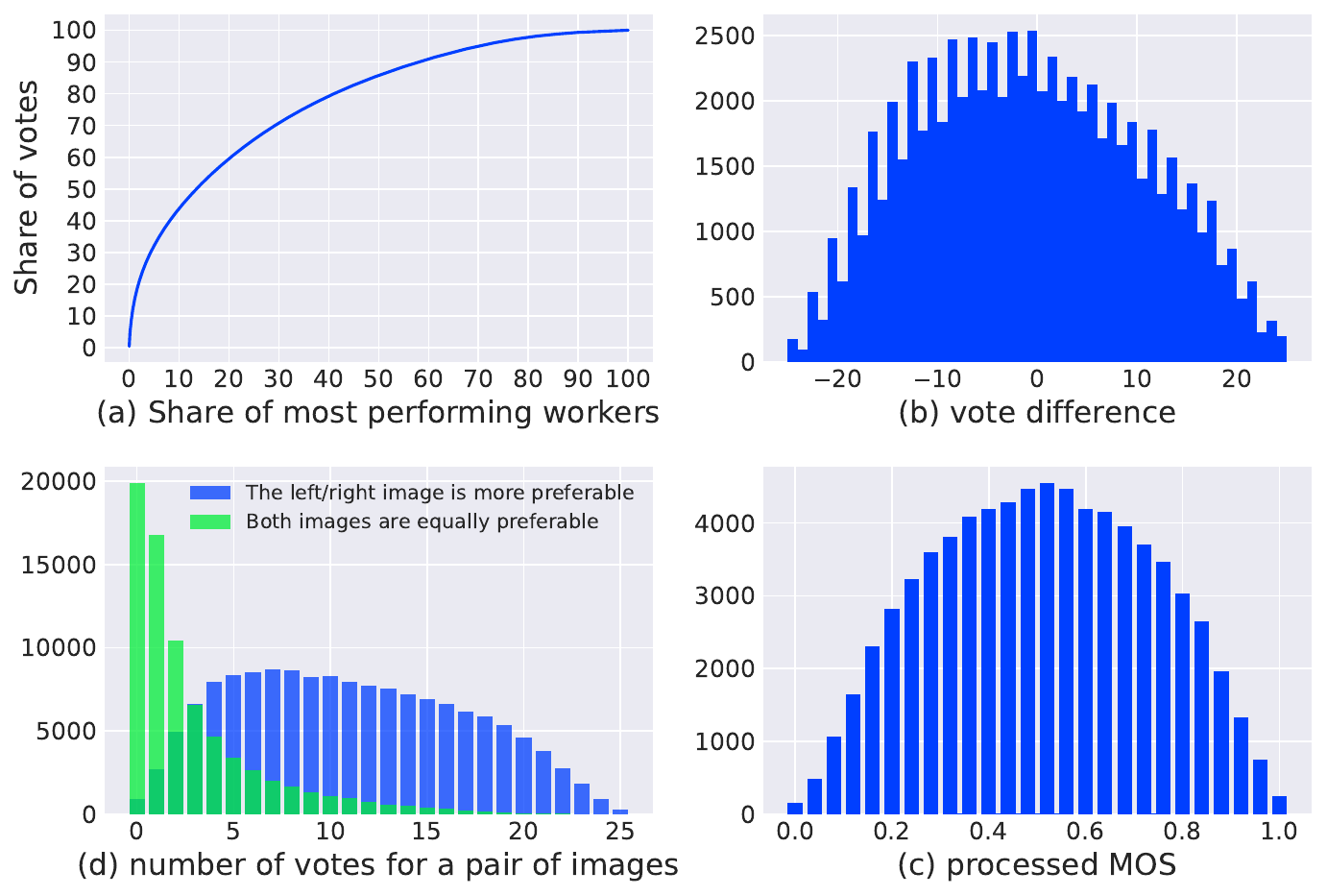}
    \caption{Analysis of DEAR: (a) Distribution of annotation workload. (b) The vote difference distribution. Vote difference is obtained by subtracting votes for left and right image, while votes that claim images look the same are not used. The distribution looks inconsistent due to distribution of votes for ``Both images are equally preferable''. (c) Processed Difference Mean Opinion Score (DMOS) distribution. DMOS is calculated via Equation~\ref{eq:dmos} that smooths the final distribution. (d) The distribution of votes per each image pair in experiment. Votes for ``Both images are equally preferable'' are not distributed similar to other types. Instead, this category gives bias, making processing more complex.}
    \label{fig:votes_hist}
\end{figure}

Overall, 13,648 people annotated the dataset, completing 2.5 million tasks, including those in the test procedure. 
In particular, the distribution of votes follows a power law: the most active 20\% of users generated 60\% of the data, while the top 30\% generated 70\%. While this has no direct and immediate consequence on the quality of the dataset, it is still interesting to mention.
For a detailed view, see Figure~\ref{fig:votes_hist}a.

The distribution of votes is shown in Figure~\ref{fig:votes_hist}d. 
It can be seen that vote ``Both images are equally preferable'' reminds of the exponential distribution.
Due to even numbers being more probable here, the probability of total number of votes that determines quality (``The left image is preferable'' and ``The right image is preferable'') is more probable to be odd, since subtracting an even number from $25$ results in an odd number. 
Therefore, the distribution becomes inconsistent and non-trivial to process, see Figure~\ref{fig:votes_hist}b.
To deal with this issue, we estimated scores by using Equation~\ref{eq:dmos}, see Figure~\ref{fig:votes_hist}c,
\begin{equation}\label{eq:dmos}
    DMOS = L - R + B \mod 2, 
\end{equation}
where $L, R, B$ are the numbers for votes for left, right, and both images, respectively.

\begin{figure}[ht]
    \centering
    \includegraphics[width=0.6\linewidth]{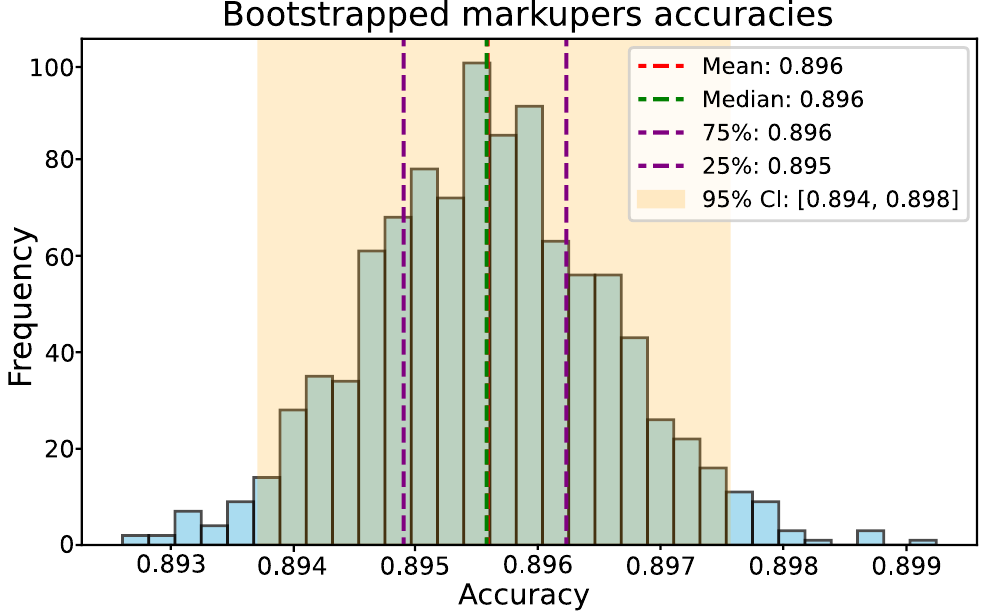}
    \caption{Histogram of maximum accuracy estimation across 1000 resampled datasets. The distribution illustrates the upper bound of performance achievable by a model that perfectly replicates the voting behavior of 25 human evaluators.}
    \label{fig:bootstrap_accuracy}
\end{figure}

To assess the reliability and potential ceiling of model performance, we estimated the maximum achievable accuracy through bootstrapping.
At each iteration, votes were randomly drawn from the original annotation pool to construct a new dataset, subsequently binarized into preferred style labels. 
The agreement between the resampled dataset and the original annotations was then computed as accuracy. Repeating this process 1000 times produced the histogram shown in Figure \ref{fig:bootstrap_accuracy}.
The distribution is centered at a mean accuracy of 0.896 with confidence interval of 0.004, indicating that even a model capable of flawlessly reproducing the preferences of all 25 evaluators cannot exceed this score.
This value therefore represents a practical upper bound for predictive performance on the EAR task.

\subsection{Possible Tasks}
Our dataset enables a wide range of research directions related to aesthetic perception and style evaluation.
Below, we outline several potential tasks:

\begin{enumerate}
\item \textbf{Analysis of Human Aesthetic Biases}:
Investigate how style preferences vary across different image categories (e.g., landscapes, portraits, food), color distributions, and compositional patterns.

\item \textbf{Aesthetic Preference Prediction}:  
Train a model to predict which of two retouched versions of an image users would prefer, given only the image content and/or style features.  

\item \textbf{Style Ranking}:  
Predict a ranking of user preferences for multiple styles applied to the same original image, instead of just pairwise preference prediction.  

\item \textbf{Personalized Aesthetic Preference Prediction}:  
Leverage evaluator identifiers to train models that predict the preferred style of a specific user.  
This task allows studying personalization and modeling individual aesthetic profiles.  

\item \textbf{Style Generation}:  
Explore generative models capable of producing new rendered styles that maximize predicted user preference.  

\item \textbf{Aesthetics Evaluation Benchmarking}:  
Use the dataset to evaluate and benchmark aesthetics evaluation models, style evaluation methods, and fine-tuned foundation models (e.g., CLIP).  
This is closely aligned with the current work presented in this paper.

\end{enumerate}

Our dataset consists of 5,000 images spanning a diverse set of vernacular subjects.
No explicit selection criteria were provided for image collection.
Each image was rendered by five knowledgeable photography students, i.e., renderers using Adobe Lightroom in 2011, though no information is available regarding their relative skill levels or aesthetic preferences.

These characteristics naturally lead to two complementary areas of inquiry:
(1) understanding viewer preferences and the factors that drive them, and
(2) analyzing the original rendering process and its influence on perceived quality.

In the following sections, we focus on four key tasks:
\emph{viewer preference analysis},
\emph{rendering preference prediction},
\emph{personalized rendering preference prediction},
and \emph{rendering evaluation benchmarking}.

%% file: Analysis_of_preferences.tex
\section{Analysis of preferences}\label{sec:preferences}
\begin{figure}[ht!]
    \centering
    \includegraphics[width=0.65\linewidth]{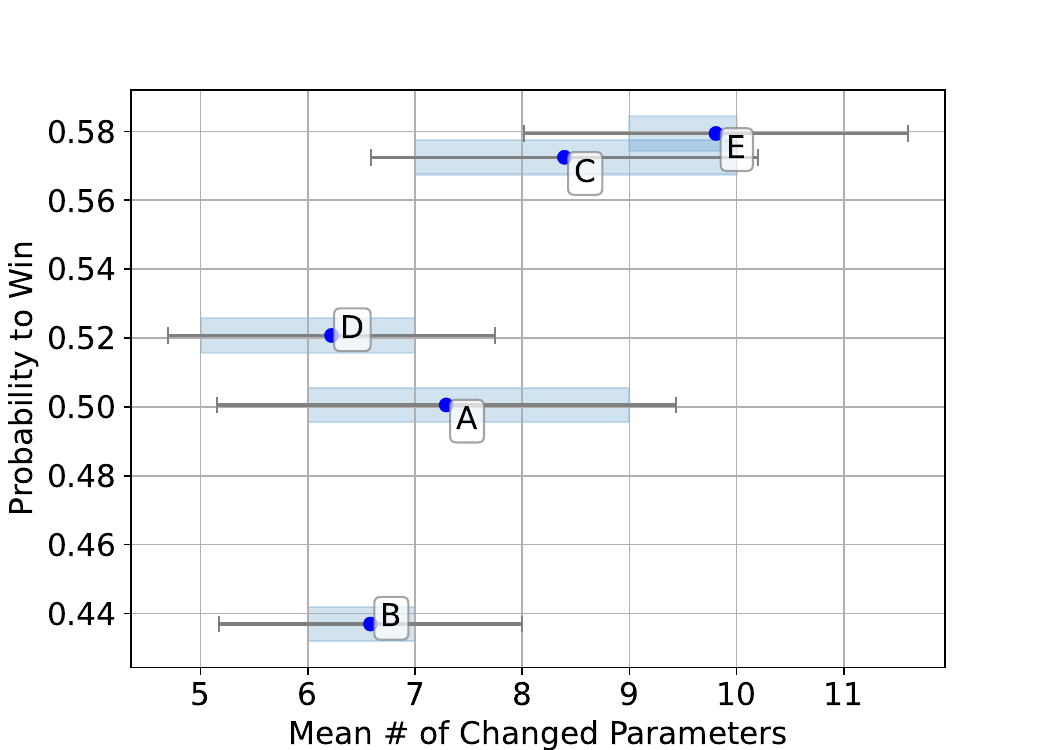}
    \caption{Analysis of used settings impact of photographers to percieved quality. Gray line represent SD, blue dot is the mean and the blue region displays $0.25$ to $0.75$ quantile. Pearson correlation is 0.65.}
    \label{fig:impact}
\end{figure}

The obtained data gives us opportunity to analyze preferences of evaluators for different categories of scenes and different styles, see Figure~\ref{fig:impact}.

A moderate correlation was observed between the mean number of parameters changed by the renderer (i.e., Expert in FiveK terminology) and the probability of win in comparison with others.
Renderer E, who had the highest mean number of altered parameters (11.5 ± 2.44), also yielded the highest win probability (57.9\%).
Conversely, renderer B, with the second-lowest mean number of changes (8.4 ± 2.25), resulted in the second-lowest win rate (43.7\%).

However, this relationship was moderated by the efficiency of the changes.
Notably, renderer D achieved a high win probability (52.1\%) with the lowest number of mean changes (7.8 ± 1.92), indicating a highly efficient parameter modification strategy.
Furthermore, while renderers A and C had an identical mean number of changes (10.6), they exhibited a substantial difference in performance (win probability of 50.1\% and 57.2\%, respectively).
This discrepancy underscores that the quality and type of parameter changes are critical factors that influence the outcome of image editing, beyond the sheer number of modifications.

The low and similar standard deviations across all renderers indicate that the application of each strategy was consistent and reproducible.

\begin{figure}[ht!]
    \centering
    \includegraphics[width=0.85\linewidth]{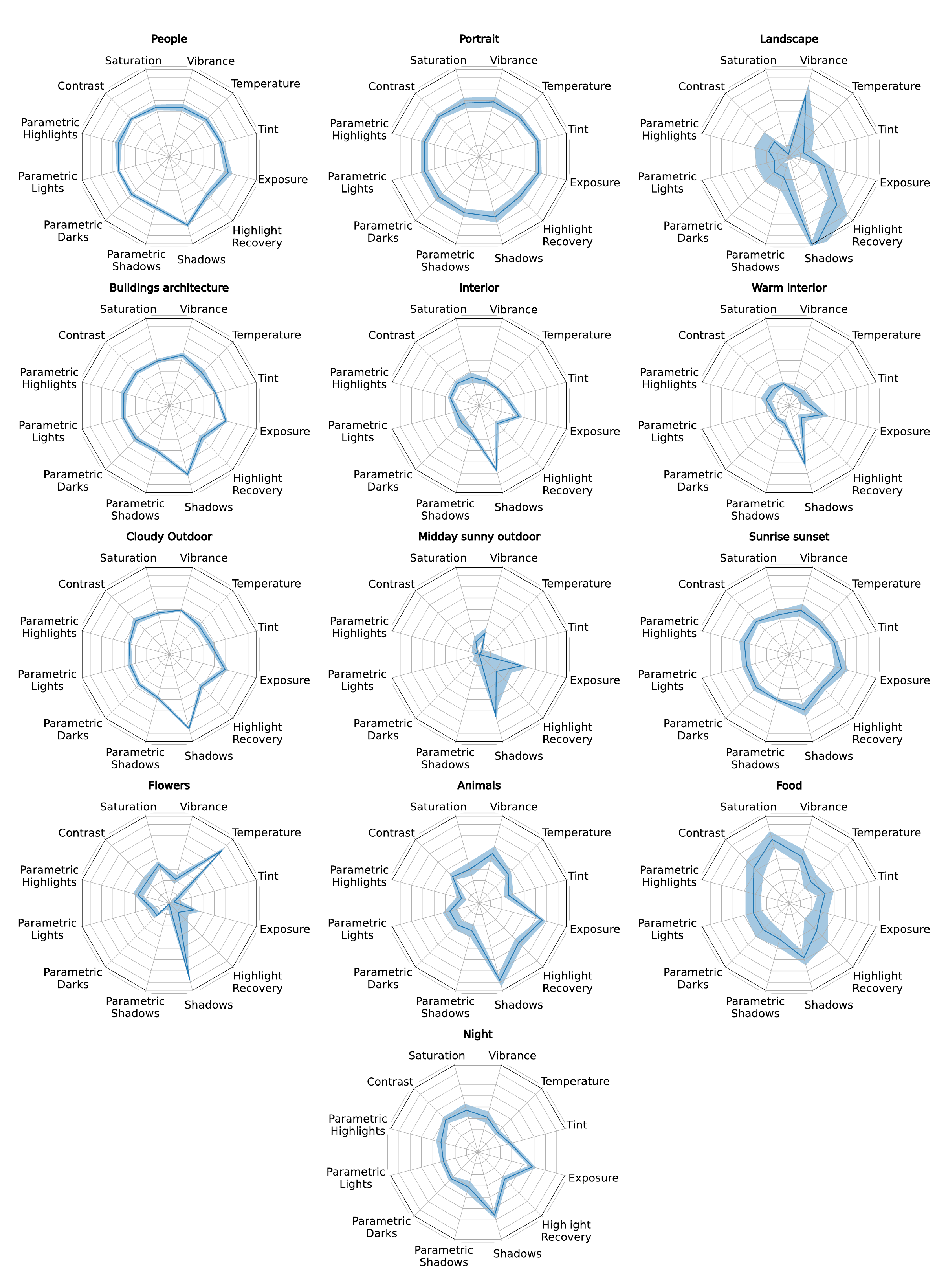}
    \caption{Radar charts of importance of main settings in Adobe Lightroom for different content groups. There is visible difference which can show that preferred style depends on content. Blue line is mean value and blue area is 0.25 and 0.75 quartile.}
    \label{fig:radrarchart}
\end{figure}

For our dataset we calculated the importance of settings for different categories to verify that the preferred style depends on content, see Figure~\ref{fig:radrarchart}.
We selected the most promising settings based on the advice of professional photographer Michael Freeman and trained linear regression for each setting.
As measure of importance we used accuracy with obtained DMOS.
For better visibility, we calculated maximum and minimum value among all categories and subtracted minimum and divided by maximum. 

From the obtained charts we can clearly see that different scene types are described with preferred setting types differently.
Moreover, for example, ``People'' and ``Portrait'' categories look similar, which makes sense, because these categories have close content fulfillment.   

%% file: quality_prediction.tex
\section{Rendering preferences prediction}
\label{sec:prediction}

There are numerous existing methods addressing the IQA task.
However, none of them were specifically designed for EAR.
In this section, we analyze state-of-the-art IQA models and train new models tailored for EAR on the proposed DEAR.

We've selected several of the most promising NR-IQA methods representing both deep learning–based and statistical approaches.
Since NR-IQA models operate on a single image at a time, and our dataset is pairwise, we first computed quality scores for each image independently and then trained a linear regression model on these scores.

For evaluation, we allocated approximately 25\% of the images to the test set.
The test set comprised $n = 1,283$ scenes, resulting in $N = 1,283 \times 15 = 19,245$ image pairs.
Some images caused invalid preprocessing issues, thus we filtered out 250 image pairs, resulting in $18,995$ image pairs. 

We adopt the following notation: the DMOS value for an image pair $i$ and $j$ within scene $s$ is denoted as $p^s_{ij}$, with the corresponding predicted score given by $\hat{p}^s_{ij}$. The aggregated pairwise score obtained via the Bradley–Terry procedure~\cite{bradley1952rank} is denoted as $q^s_i$, with its predicted counterpart $\hat{q}^s_i$.

We then evaluated performance using the following metrics:
\begin{itemize}
    \item \textbf{Accuracy}: 
    \begin{equation}
        \frac{1}{N}\displaystyle\sum_{s,i,j, i\neq j} \mathbf{1}_{p^s_{ij}>0.5}(\hat{p}^s_{ij} > 0.5) + \mathbf{1}_{p^s_{ij}\leq 0.5}(\hat{p}^s_{ij}\leq 0.5)
    \end{equation}
    \item \textbf{Top-1 accuracy}:
    \begin{equation}
        \frac{1}{n} \sum_s\mathbf{1}_{\underset{i}{\mathrm{argmax}}(q^s_{i})}(\underset{i}{\mathrm{argmax}}(\hat{q}^s_{i}))
    \end{equation}
    \item \textbf{Spearman correlation}:
    \begin{equation}
        \frac{1}{n}\sum_s\rho[R_i[q^s_i], R_i[\hat{q}^s_i]], 
    \end{equation}
    where $R_i[\cdot]$ is ranking operation for list of scores for $i$ index.
    \item \textbf{Kendall correlation}:
    \begin{equation}
        \frac{1}{n}\sum_s\tau[R_i[q^s_i], R_i[\hat{q}^s_i]] 
    \end{equation}
\end{itemize}

\subsection{Aesthetic preference prediction}

In the first block in Table~\ref{tab:metrics_for_SSAD} there are results for evaluating total variation~(TV) and pretrained models: BRISQUE\cite{Mittal2012}, PIQE~\cite{pique}, ARNIQA~\cite{arniqa} CLIP IQA\cite{Wang2023}, LARIQA~\cite{lariqa} and LIQE~\cite{Zhang2023}.
It can be seen that simple TV performs comparable with foundation-based models like LIQE.
This means that current IQA models are not applicable in an EAR task without finetuning on this data.

For the MIT-Adobe FiveK dataset, metadata describing the applied rendering settings (e.g., saturation, contrast, color temperature) is available.
This allows for score prediction based solely on these settings.
Additionally, we extracted several handcrafted features directly from the images, including standard deviation, lightness, saturation, hue, as well as $u$ and $v$ components from the UV color space.
All features were normalized to the [0, 1] range.
Moreover, due to similarity of the TV method to standard deviation (SD), we used linear regression for SD only.
The results for this experiment are shown in the second block of Table~\ref{tab:metrics_for_SSAD}.
It is worth noting that this simple combination of image settings and handcrafted features outperforms all conventional IQA metrics.
Even using only SD as a feature yields a better result than most existing IQA methods.

Finally, we trained several models specifically for the score prediction task.
Because we work with image pairs, we explored two strategies: the Siamese approach (``$\text{model}_\text{siam}$''), i.e., forwarding both images independently through the base model and combining the outputs with a linear regression, and ConCat approach (``$\text{model}_\text{cc}$''), i.e., concatenating the two images along the channel dimension (resulting in a tensor with a depth dimension equal to 6), and directly predicting the score.
It is worth noting that for the CLIP model we freeze the vision transformer (ViT) weights training linear regression only. 
The results are shown in the third block of Table~\ref{tab:metrics_for_SSAD}.
It can be seen that the Siamese approach is strictly worse than stacking two images in one tensor. 
Otherwise, ConCat approach limits usage of foundation models due to the specificity of the hidden representation.

We further enhanced these models by incorporating metadata and the handcrafted features from the previous experiment; this extended setup is presented in the fourth block of Table~\ref{tab:metrics_for_SSAD}.
For the CLIP model we used pretrained CLIP VIT-B/32 to generate the embeddings.
Using these embeddings and the extracted features, we trained a bilinear regression model, where the first argument consisted of CLIP embedding and the second consisted of the extracted features.
\textit{Note:} this model has a significant downside, as it implies that the best style for a given scene consists of min-max slider values.
Despite the added information, these models yield even worse results than the image-based approaches.
A range of illustration demonstrating quality evaluation for diverse rendering can be found in Supplementary Materials.
Detailed technical information on training and testing procedures are available in project GitHub repository.

\begin{table*}[t]
    \centering
    \caption{Evaluated metrics for different methods over the pairwise comparison task. The \textbf{best-performing} models for each metric \textbf{in block} is highlighted in \textbf{bold}. The \textcolor{teal}{best-performing} and \textcolor{purple}{second-best} models for each metric are highlighted in \textcolor{teal}{teal} and \textcolor{purple}{purple}, respectively. }
\begin{tabular}{|c|l|cccc|}\hline\hline

 & & Accuracy $\uparrow$ & Spearman $\uparrow$ & Kendall $\uparrow$ & Top 1 accuracy $\uparrow$ \\\hline

\multirow{7}{*}{\rotatebox[origin=c]{90}{Pretrained}}
 & TV & \textbf{0.73} & \textbf{0.56} & 0.47 & \textbf{0.50} \\
 & BRISQUE & 0.64 & 0.35 & 0.29 & 0.35 \\
 & PIQE & 0.58 & 0.20 & 0.17 & 0.27 \\
 & ARNIQA & 0.65 & 0.39 & 0.32 & 0.38 \\
 & CLIP IQA & 0.68 & 0.45 & 0.38 & 0.43 \\
 & LARIQA & 0.63 & 0.33 & 0.27 & 0.32 \\
 & LIQE & \textbf{0.73} & \textbf{0.56} & \textbf{0.48} & 0.48 \\
\hline
\multirow{3}{*}{\rotatebox[origin=c]{90}{\shortstack{Feature\space\\based}}} 
 & Linear Regression & \textbf{0.76} & \textbf{0.62} & \textbf{0.53} & \textbf{0.52} \\
 & Random Forest & 0.74 & 0.60 & 0.51 & 0.50 \\
 & SD Linear Regression & 0.71 & 0.49 & 0.42 & 0.44 \\
\hline
\multirow{9}{*}{\rotatebox[origin=c]{90}{\shortstack{Image\\based}}} 
 & Siamese CLIP ViT-B/32 & 0.78 & 0.69 & 0.58 & 0.53 \\
 & $\text{Resnet152}_\text{cc}$ & \textcolor{teal}{\textbf{0.81}} & \textcolor{purple}{0.74} & \textcolor{purple}{0.64} & \textcolor{purple}{0.57} \\
 & $\text{Resnet50}_{cc}$ & \textcolor{teal}{\textbf{0.81}} & \textcolor{purple}{0.74} & \textcolor{purple}{0.64} & \textcolor{purple}{0.57} \\
 & $\text{MobileNetV3(L)}_\text{cc}$ & \textcolor{teal}{\textbf{0.81}} & \textcolor{teal}{\textbf{0.75}} & \textcolor{teal}{\textbf{0.65}} & \textcolor{teal}{\textbf{0.58}} \\
 & $\text{MobileNetV2}_\text{cc}$ & \textcolor{purple}{0.80} & 0.73 & 0.62 & 0.56 \\
 & $\text{Resnet152}_\text{siam}$ & 0.79 & 0.71 & 0.61 & 0.53 \\
 & $\text{Resnet50}_\text{siam}$ & 0.79 & 0.70 & 0.60 & 0.52 \\
 & $\text{MobileNetV3(L)}_\text{siam}$ & 0.79 & 0.71 & 0.61 & 0.55 \\
& $\text{MobileNetV2}_\text{siam}$ & 0.78 & 0.68 & 0.58 & 0.53 \\

 \hline

 \multirow{4}{*}{\rotatebox[origin=c]{90}{\shortstack{Image\\Feature\\based}}} 
 & $\text{Resnet152}_\text{cc}$ & \textcolor{purple}{0.80} & \textcolor{purple}{0.74} & \textcolor{purple}{\textbf{0.64}} & \textcolor{purple}{\textbf{0.57}} \\
 & $\text{MobileNetV3(L)}_\text{cc}$ & \textcolor{teal}{\textbf{0.81}} & \textcolor{teal}{\textbf{0.75}} & \textcolor{purple}{\textbf{0.64}} & 0.56 \\
 & $\text{Resnet152}_\text{siam}$ & 0.79 & 0.71 & 0.61 & 0.54 \\
 & $\text{MobileNetV3(L)}_\text{siam}$ & \textcolor{purple}{0.80} & 0.72 & 0.61 & 0.55 \\

 \hline\hline

\end{tabular}

    \label{tab:metrics_for_SSAD}
\end{table*}

\subsection{Personalized EAR}
In the previous section, we analyzed the aggregated preferences of all voters.
In this experiment, we focus on a single voter: the most active one.
This individual voter provided 7,678 votes, covering 595 scenes in which he/she voted on every available style at least once.
Given the relatively small size of this subset, we restricted our evaluation to simple models based on handcrafted image features and CLIP embeddings.
The results are reported in Table~\ref{tab:metrics_for_person}.
It can be seen that individual prediction performs worse than similar methods in Table~\ref{tab:metrics_for_SSAD} except for top-1 accuracy.
This may be the result of a well-known issue, where individuals are confident about the best and the worst examples and are indifferent to anything in between~\cite{parducci1965category,tversky1974judgment}. 

\begin{table*}[t]
    \centering
    \caption{Evaluated metrics for different methods obtained on the pairwise comparison task for the personalized EAR.The \textbf{best-performing} model for each metric is highlighted in \textbf{bold}.}
\begin{tabular}{|l|cccc|}
\hline\hline
 & Accuracy $\uparrow$ & Spearman $\uparrow$ & Kendall $\uparrow$ & Top 1 accuracy $\uparrow$ \\
\hline
TV & 0.67 & 0.35 & 0.35 & 0.58 \\
BRISQUE & 0.68 & 0.36 & 0.36 & 0.58 \\
PIQE & 0.67 & 0.34 & 0.33 & 0.57 \\
ARNIQA & 0.66 & 0.32 & 0.31 & 0.56 \\
CLIP IQA & 0.66 & 0.32 & 0.31 & 0.56 \\
LARIQA & 0.66 & 0.32 & 0.31 & 0.56 \\
LIQE & 0.68 & 0.36 & 0.35 & 0.57 \\
\hline
SD Linear Regression & 0.67 & 0.36 & 0.36 & 0.58 \\
Linear Regression & \textbf{0.71} & \textbf{0.43} & \textbf{0.39} & \textbf{0.59    } \\
Random Forest & 0.70 & 0.38 & 0.35 & 0.56 \\
Siamese CLIP ViT-B/32 & 0.70 & 0.26 & 0.24 & 0.29 \\
\hline\hline
\end{tabular}

    \label{tab:metrics_for_person}
\end{table*}

\subsection{Comparison of distortion-based IQA on an EAR task}

A final experiment was conducted to compare IQA tasks for distortions with EAR.
The key idea was to evaluate a set IQA models and benchmark datasets, and then cross-validate their performance with DEAR and our proposed Model for Evaluating the Aesthetics of Rendering (MEAR).

For this comparison, we selected $\text{MobileNetV3(L)}_\text{cc}$ as the representative implementation of MEAR, since it demonstrated the strongest performance across most metrics on the DEAR benchmark.

Because our model operates on image pairs, we adapted existing IQA datasets by using distorted–reference pairs as inputs.
As the evaluation metric, we computed the Spearman correlation between predicted scores and subjective ratings for all images associated with the same reference, and then averaged these correlations across reference scenes.
The results of this experiment are summarized in Table~\ref{tab:benchmark}.

It can be observed that all existing models, with the exception of MEAR, achieve relatively low correlations on the DEAR benchmark.
At the same time, MEAR exhibits comparatively weaker performance on distortion-oriented IQA tasks, while providing substantially higher correlation for EAR.
These results indicate that EAR is fundamentally different from traditional distortion-based IQA and that each task benefits from models specifically trained to address its unique objectives.

\begin{table*}[t]
    \centering
    \caption{Spearman correlation for different models on different datasets. MEAR is the best model, trained on DEAR. The \textbf{best-performing} model for each dataset is highlighted in \textbf{bold}.}
    \footnotesize
    \setlength{\tabcolsep}{5.9pt} 
    \begin{tabular}{|l|ccccccc|c|}
    \hline\hline
    \diagbox[]{dataset}{model} & TV & BRISQUE & PIQE & ARNIQA & CLIP-IQA & LIQE & LARIQA & \textbf{MEAR} \\
    \hline
    VCL@FER     & 0.14& \textbf{0.90}& 0.79& \textbf{0.90}& 0.56& \textbf{0.90}& 0.66& 0.49 \\
    LIVE        & 0.30& \textbf{0.96}& 0.85& 0.91& 0.73& 0.92& 0.72& 0.26 \\
    CSIQ        & 0.27& 0.55& 0.54& 0.89& 0.87& \textbf{0.93}& 0.61& 0.12 \\
    TID2013     & 0.02& 0.39& 0.36& \textbf{0.75}& 0.61& 0.74& 0.38& 0.01 \\
    KADID-10k   & 0.12& 0.35& 0.25& 0.93& 0.56& \textbf{0.79}& 0.49& 0.19 \\
    PDAP-HDDS   & 0.16& 0.41& 0.32& \textbf{0.64}& 0.42& 0.62& 0.46& 0.28 \\\hline
    \textbf{DEAR} & 0.61& 0.34& 0.07& 0.43& 0.53& 0.49& 0.29&  \textbf{0.78} \\
    \hline\hline
    \end{tabular}
    \label{tab:benchmark}
\end{table*}

%% file: discussion.tex
\section{Discussion}
\label{sec:discussion}

As demonstrated in the previous sections, the DEAR introduces a novel and challenging problem setting within the field of IQA: namely, EAR.
Existing IQA methods exhibit substantially lower performance on DEAR compared to their results on traditional IQA datasets, as well as relative to the maximum achievable accuracy.
Even models trained specifically for the EAR task reach only 0.81 accuracy, whereas the empirically found practical upper bound appears to be about 0.89.
This gap suggests that predicting the quality of rendering remains a difficult problem, likely due to the subjective and fine-grained nature of aesthetics judgments.

One factor contributing to the difficulty of the task is the prevalence of image pairs with preference scores near 0.5, indicating ambiguous or subtle differences between styles.
These ``hard'' pairs reduce the separability of classes and limit the performance of both classical and deep learning–based methods.

Although metadata describing the applied rendering parameters (e.g., saturation, contrast, color temperature) have information useful for prediction to some extent, its use alone does not yield state-of-the-art performance.
Moreover, simply concatenating this metadata with deep learning embeddings does not significantly improve the results.
A promising future direction may involve explicitly modeling the effect of each retouching parameter --- potentially through submodules that approximate the transformations applied by Adobe Lightroom --- and integrating these parameter-conditioned transformations into the network.
Such an approach, however, would require partial reverse engineering of Adobe's rendering pipeline or learning differentiable approximations of these adjustments.

An important advantage of DEAR is its potential for extensibility.
The dataset can be expanded by collecting additional votes, which would reduce noise and improve the stability of DMOS.
In particular, increasing the number of evaluators per image would help mitigate the impact of individual biases and provide a more robust representation of collective aesthetic preference.
Such expansions would make DEAR even more valuable as a benchmark for fine-grained aesthetic prediction.

Despite its advantages, DEAR has several limitations.
The current version contains only six rendering styles, which represent a limited portion of the possible parameter space.
This excludes many extreme or intentionally poor combinations of parameters (e.g., overly saturated or severely underexposed images) that could help models learn a wider range of aesthetic judgments.
Future versions of the dataset could incorporate a richer and more diverse set of retouching styles to broaden its coverage and challenge models to generalize across a wider aesthetic spectrum.

Beyond its immediate utility for benchmarking, DEAR opens opportunities for several research directions.
It can serve as a benchmark for developing models capable of learning subtle perceptual differences, as well as for studying human aesthetic biases across image categories and styles.
Furthermore, the availability of evaluator identifiers allows investigation of personalization and user-specific preference modeling --- an area of growing importance in content curation, recommendation systems, and generative AI.
Finally, the dataset provides a bridge between traditional IQA research and the broader study of image aesthetics, motivating the development of models that move beyond distortion-based quality prediction toward understanding visual appeal.

%% file: concl.tex
\section{Conclusions}
\label{sec:conclusions}

In this work, we introduced the DEAR --- a novel benchmark for studying aesthetic preferences prediction and style evaluation through the EAR task.
DEAR provides pairwise preference annotations for 5,000 images, each rendered into multiple styles, making it uniquely suited for fine-grained analysis of human aesthetic judgments.
Through extensive experiments, we demonstrated that existing IQA methods perform poorly on the EAR task compared to their results on standard quality-assessment datasets, highlighting the need for new approaches that explicitly model aesthetic factors.
Even purpose-trained models achieved only 0.81 accuracy, leaving a substantial gap from the achievable upper bound of 0.89.

We further explored the predictive power of image metadata and handcrafted features, finding that they offer useful but limited information, and we showed that integrating them with deep learning models does not trivially improve performance.
These findings emphasize the complexity of aesthetic perception and suggest that future models may need to incorporate parameter-aware or differentiable style modules to better capture the underlying retouch transformations.

DEAR not only provides a challenging benchmark, but also opens several promising research directions, including personalized preference prediction, style ranking, and generative modeling of preferred styles.
As a dataset, DEAR can be extended with additional votes and new styles, improving the robustness of mean opinion scores and expanding the diversity of aesthetic scenarios.
We hope that DEAR will foster the development of next-generation aesthetic prediction models and serve as a common ground for comparing approaches at the intersection of IQA, computational aesthetics, and human-AI co-creation.

Future work will include generating new styles that cover parameter field more broadly in order to obtain more generalized model that can be used as loss for automatically creating the most preferable styles.
Moreover, generating parameter-conditioned transformations and integrating them into the network might reduce the gap with respect to the maximum achievable performance.

%% file: supplementary.tex









\section{Supplementary for DEAR: Dataset for Evaluating the Aesthetics of Rendering}
\subsection{BLIP}
To study the dependence of the importance of sliders on the content of the scene, we marked up scenes from MIT-Adobe Fivek using bootstrapping language-image pre-training (BLIP) from  Salesforce\footnote{\href{https://huggingface.co/Salesforce/blip-vqa-base}{https://huggingface.co/Salesforce/blip-vqa-base}}. The markup prompts were as follows:
\begin{itemize}
    \item \textbf{People}: Is it some people in a casual setting?
    \item \textbf{Portrait}: Is it close-up photo of a person's face?
    \item \textbf{Landscape}: Is it natural scenery: mountains, fields, water, desert, canyon?
    \item \textbf{Buildings architecture}: Is it exterior view of buildings, architectural structures?
    \item \textbf{Interior}: Is it indoor room, interior items?
    \item \textbf{Warm interior}: Is it indoor scene with warm lighting?
    \item \textbf{Cloudy Outdoor}: Is it overcast outdoor scene with diffuse lighting and soft shadows?
    \item \textbf{Midday sunny outdoor}: Is it bright outdoor scene with harsh shadows and high contrast?
    \item \textbf{Sunrise sunset}: Is it sky with vibrant colors near the horizon during dawn, dusk?
    \item \textbf{Flowers}: Is it close-up of blooming flowers or floral arrangements?
    \item \textbf{Animals}: Is it photo of an animal: pet, mammal, bird, fish, insect?
    \item \textbf{Food}: Is it appetizing close-up of prepared meal?
    \item \textbf{Night}: Is it low-light scene taken after dark with artificial lighting?
\end{itemize}
We got ``yes'' or ``no'' answers for each rendering of a scene and then selected the final answer for a scene by the majority of votes.
For the accuracy estimation, we manually labeled 100 scenes. The computed metrics are reported in Table~\ref{tab:BLIP_accuracy}.

\begin{table*}[t]
\centering
\caption{Accuracy and other metrics for the BLIP markup.}
\begin{tabular}{l|c|c|c|c|c|c|c|c}
 & Accuracy & F1 Score & Precision & Recall & TP,\% & FP,\%& FN,\% & TN,\% \\
\hline
People & 0.92 & 0.87 & 0.84 & 0.90 & 26  & 5  & 3  & 66  \\
Portrait & 0.97 & 0.90 & 0.93 & 0.88 & 14  & 1  & 2  & 83  \\
Landscape & 0.95 & 0.78 & 1.00 & 0.64 & 9  & 0  & 5  & 86  \\
Buildings architecture & 0.96 & 0.95 & 0.92 & 0.97 & 36  & 3  & 1  & 60  \\
Interior & 0.99 & 0.98 & 1.00 & 0.96 & 25  & 0  & 1  & 74  \\
Warm interior & 0.96 & 0.90 & 0.82 & 1.00 & 18  & 4  & 0  & 78  \\
Cloudy Outdoor & 0.91 & 0.91 & 0.85 & 0.98 & 44  & 8  & 1  & 47  \\
Midday sunny outdoor & 0.88 & 0.57 & 0.89 & 0.42 & 8  & 1  & 11  & 80  \\
Sunrise sunset & 0.98 & 0.93 & 0.87 & 1.00 & 13  & 2  & 0  & 85  \\
Flowers & 0.99 & 0.95 & 1.00 & 0.90 & 9  & 0  & 1  & 90  \\
Animals & 0.93 & 0.70 & 0.73 & 0.67 & 8  & 3  & 4  & 85  \\
Food & 0.99 & 0.94 & 1.00 & 0.89 & 8  & 0  & 1  & 91  \\
Night & 0.93 & 0.84 & 0.72 & 1.00 & 18  & 7  & 0  & 75  \\
\end{tabular}
\label{tab:BLIP_accuracy}
\end{table*}

\subsection{Expert markup analysis}
\begin{figure}[!ht]
    \centering
    \includegraphics[width=\linewidth]{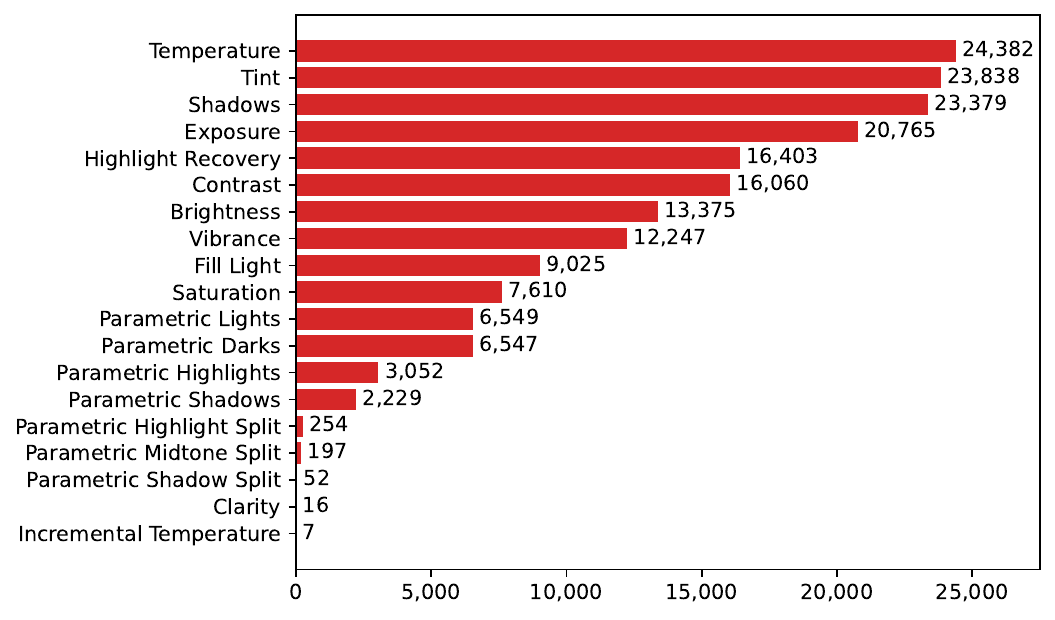}
    \caption{Number of changes for most used settings in MIT-Adobe Fivek.}
    \label{fig:slider_usage}
\end{figure}

In Figure~\ref{fig:slider_usage} we report the estimated number of settings used in MIT-Adobe FiveK. 
Despite having a large variety of sliders to change, approximately 90, the renderers only changed a slight amount of them.

\subsection{Examples of model prediction}
\begin{figure}[!ht]
    \centering
    \includegraphics[width=0.95\linewidth]{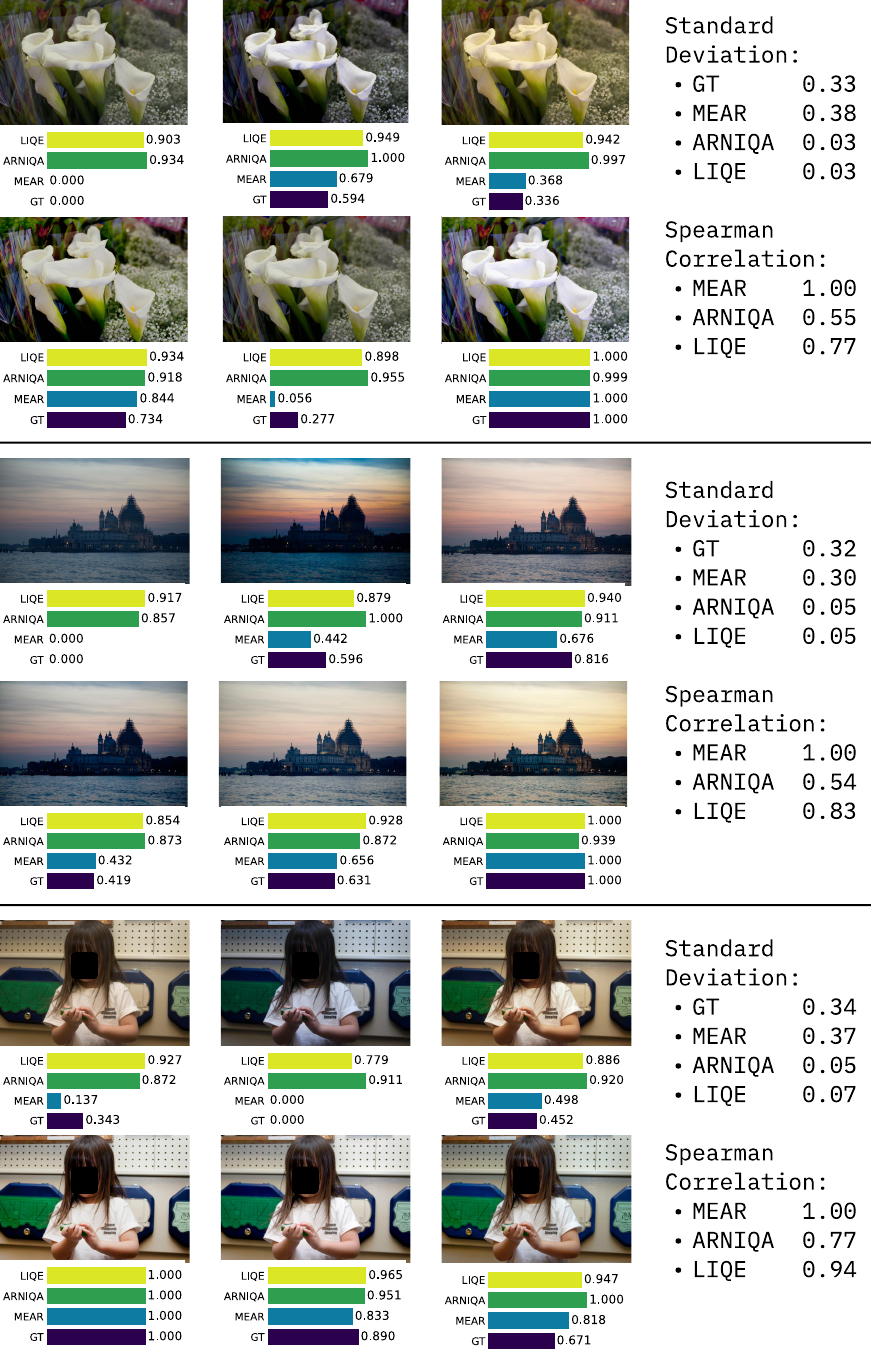}
    \caption{Examples of images. Our model shows visibly more attention to rendering preferences than existing methods.}
    \label{fig:examples}
\end{figure}

In Figure \ref{fig:examples} we illustrate some examples of scenes where significantly different levels of aesthetic of rendering are correctly estimated with our model MEAR, whereas other state-of-the-art models (i.e. LIQE and ARNIQA) judge them to be more or less aesthetically equivalent.

%% file: sn-bibliography.bib
@article{zhai2020perceptual,
  title="Perceptual image quality assessment: a survey",
  author="Zhai, Guangtao and Min, Xiongkuo",
  journal="Science China Information Sciences",
  volume="63",
  pages="211301",
  year="2020",
  publisher="Springer",
  doi={10.1007/s11432-019-2757-1}
}

@misc{ershov2024reliability,
  title={{R}eliability and {S}tability of {M}ean {O}pinion {S}core for {I}mage {A}esthetic {Q}uality {A}ssessment {O}btained {T}hrough {C}rowdsourcing},
  author={Ershov, Egor and Panshin, Artyom and Ermakov, Ivan and Bani{\'{c}'}, Nikola and Savchik, Alex and Bianco, Simone},

  volume={365},

  year={2024},
  note=  {Paper presented at the 19th International Joint Conference on Computer Vision, Imaging and Computer Graphics Theory and Applications, Rome, Italy, 27--29 February 2024},
  doi = {10.5220/0012462000003660}
}

@article{Zari2012,
   author = {{Andela, Zari{\'{c}}} and {Nenad, Tatalovi{\'{c}}} and {Nikolina, Brajkovi{\'{c}}} and {Hrvoje, Hlevnjak} and {Matej, Lon{\v{c}}ari{\'{c}}} and {Emil, Dumi{\'{c}}} and {Sonja, Grgi{\'{c}}}},
   doi = {10.7305/automatika.53-4.241},

   number = {4},
   journal = {Automatika},
   pages = {344--354},
   title = {{V}{C}{L}@{F}{E}{R} {I}mage {Q}uality {A}ssessment {D}atabase},
   volume = {53},
   year = {2012}
}

@article{Liu2018,
   author = {Tsung, Jung Liu and Hsin, Hua Liu and Soo, Chang Pei and Kuan, Hsien Liu},
   doi = {10.1109/ACCESS.2018.2864514},

   journal = {IEEE Access},
   month = {8},
   pages = {45427-45438},
   publisher = {Institute of Electrical and Electronics Engineers Inc.},
   title = {{A} {H}igh-{D}efinition {D}iversity-{S}cene {D}atabase for {I}mage {Q}uality {A}ssessment},
   volume = {6},
   year = {2018}
}

@misc{Lin2020,
   author = {Hanhe, Lin and Vlad, Hosu and Dietmar, Saupe},
   month = {1},
   title = {{D}eep{F}{L}-{I}{Q}{A}: {W}eak {S}upervision for {D}eep {I}{Q}{A} {F}eature {L}earning},
   year = {2020},
   note = {Preprint at \url{http://arxiv.org/abs/2001.08113}}
}

@article{Hosu2019,
   author = {Vlad, Hosu and Hanhe, Lin and Tamas, Sziranyi and Dietmar, Saupe},
   doi = {10.1109/TIP.2020.2967829},
   month = {10},
   title = {{K}on{I}{Q}-10k: {A}n ecologically valid database for deep learning of blind image quality assessment},
   journal={IEEE Transactions on Image Processing},
   volume={29},
   pages={4041--4056},
   year={2020},
   publisher={IEEE}
}

@article{Li2023,
   author = {Chunyi, Li and Zicheng, Zhang and Haoning, Wu and Wei, Sun and Xiongkuo, Min and Xiaohong, Liu and Guangtao, Zhai and Weisi, Lin},
   month = {6},
   title = {{A}{G}{I}{Q}{A}-3{K}: {A}n {O}pen {D}atabase for {A}{I}-{G}enerated {I}mage {Q}uality {A}ssessment},
   doi = {10.1109/TCSVT.2023.3319020},
   journal={IEEE Transactions on Circuits and Systems for Video Technology},
   volume={34},
   number={8},
   pages={6833--6846},
   year={2023},
   publisher={IEEE}
}

@article{Ruikar2023,
AUTHOR = {Ruikar, Jayesh and Chaudhury, Saurabh},
TITLE = {{N}{I}{T}{S}-{I}{Q}{A} {D}atabase: {A} {N}ew {I}mage {Q}uality {A}ssessment {D}atabase},
JOURNAL = {Sensors},
VOLUME = {23},
YEAR = {2023},
NUMBER = {4},
pages = {2279},
URL = {https://www.mdpi.com/1424-8220/23/4/2279},
PubMedID = {36850877},
ISSN = {1424-8220},
doi = {10.3390/s23042279}
}

@article{Sun2017,
   author = {Wen, Sun and Fei, Zhou and Qingmin, Liao},
   doi = {10.1016/j.patcog.2016.07.033},
   journal = {Pattern Recognition},
   month = {1},
   pages = {153--168},
   publisher = {Elsevier Ltd},
   title = {{M}{D}{I}{D}: {A} multiply distorted image database for image quality assessment},
   volume = {61},
   year = {2017}
}

@article{Ponomarenko2015,
   author = {Ponomarenko, Nikolay and Jin, Lina and Ieremeiev, Oleg and Lukin, Vladimir and Egiazarian, Karen and Astola, Jaakko and Vozel, Benoit and Chehdi, Kacem and Carli, Marco and Battisti, Federica and others},
   doi = {10.1016/j.image.2014.10.009},
   journal = {Image Communication},
   month = {1},
   pages = {57--77},
   publisher = {Elsevier B.V.},
   title = {{I}mage database {T}{I}{D}2013: {P}eculiarities, results and perspectives},
   volume = {30},
   year = {2015}
}

@misc{Hosu2024,
  title={{U}{H}{D}-{I}{Q}{A} benchmark database: {P}ushing the boundaries of blind photo quality assessment},
  author={Hosu, Vlad and Agnolucci, Lorenzo and Wiedemann, Oliver and Iso, Daisuke and Saupe, Dietmar},
note = {Preprint at \url{https://arxiv.org/abs/2406.17472}},
  year={2024}
}

@misc{sheikh2005live,
  title={LIVE image quality assessment database release 2},
  author={Sheikh, H.R. and Wang, Z. and Cormack, L. and Bovik, A.C.},
  note={Dataset source \url{http://live.ece.utexas.edu/research/quality}},
  year={2005}
}

@article{Chandler2010,
   author = {Damon M. Chandler},
   doi = {10.1117/1.3267105},
   number = {1},
   journal = {Journal of Electronic Imaging},
   month = {1},
   pages = {011006},
   publisher = {SPIE-Intl Soc Optical Eng},
   title = {{M}ost apparent distortion: full-reference image quality assessment and the role of strategy},
   volume = {19},
   year = {2010}
}

@article{7460955,
  author={Siahaan, Ernestasia and Hanjalic, Alan and Redi, Judith},
  journal={IEEE Transactions on Multimedia}, 
  title={A Reliable Methodology to Collect Ground Truth Data of Image Aesthetic Appeal}, 
  year={2016},
  volume={18},
  number={7},
  pages={1338-1350},
  doi={10.1109/TMM.2016.2559942}}

@misc{hosu2018expertise,
  doi = {10.1109/QoMEX.2018.8463427},
  title={Expertise screening in crowdsourcing image quality},
  author={Hosu, Vlad and Lin, Hanhe and Saupe, Dietmar},
  note={Paper presented at the 2018 Tenth international conference on quality of multimedia experience (QoMEX), Cagliari, Italy, 29 May -- 01 June 2018},
  pages={1--6},
  year={2018},
}

@misc{murray2012ava,
  title={AVA: A large-scale database for aesthetic visual analysis},
  author={Murray, Naila and Marchesotti, Luca and Perronnin, Florent},
  note ={Paper presented at the 2012 IEEE 2012 IEEE Conference on Computer Vision and Pattern Recognition, Providence, RI, USA,16--21 June 2012},
  year={2012},
  doi = {10.1109/CVPR.2012.6247954}
}

@misc{wu2023human,
  title={Human preference score v2: A solid benchmark for evaluating human preferences of text-to-image synthesis},
  author={Wu, Xiaoshi and Hao, Yiming and Sun, Keqiang and Chen, Yixiong and Zhu, Feng and Zhao, Rui and Li, Hongsheng},
  note = {Preprint at \url{https://arxiv.org/abs/2306.09341}},
  year={2023}
}

@misc{kang2020eva,
  title={Eva: An explainable visual aesthetics dataset},
  author={Kang, Chen and Valenzise, Giuseppe and Dufaux, Fr{\'e}d{\'e}ric},
  doi = {10.1145/3423268.3423590},
  note ={Paper presented at the 28th ACM International Conference on Multimedia, Seattle, WA, USA, 12 -- 16 October 2020},
  year={2020}
}

@misc{kairanbay2018towards,
  title={Towards demographic-based photographic aesthetics prediction for portraitures},
  author={Kairanbay, Magzhan and See, John and Wong, Lai-Kuan},
  note ={Paper presented at the MultiMedia Modeling: 24th International Conference, MMM 2018, Bangkok, Thailand, 5--7 February 2018},
  year={2018},
  doi = {10.1007/978-3-319-73603-7\_43},
}

@misc{li2024aigiqa,
  title={{A}{i}{g}{i}{q}{a}-20k: A large database for ai-generated image quality assessment},
  author={Li, Chunyi and Kou, Tengchuan and Gao, Yixuan and Cao, Yuqin and Sun, Wei and Zhang, Zicheng and Zhou, Yingjie and Zhang, Zhichao and Zhang, Weixia and Wu, Haoning and others},
  note = {Paper presented at the 2024 IEEE Conference on Computer Vision and Pattern Recognition, Seattle WA, USA, 17--21 June 2024},
  year={2024},
  doi = {10.48550/arXiv.2404.03407}
}

@misc{yuan2023pku,
title={PKU-I2IAQA: An image-to-image quality assessment database for ai generated images},
  author={Yuan, Jiquan and Cao, Xinyan and Li, Changjin and Yang, Fanyi and Lin, Jinlong and Cao, Xixin},
  note ={Preprint at \url{https://arxiv.org/abs/2311.15556}},
  year = {2023}
}

@article{Mittal2012,
   author = {Anish, Mittal and Anush, Krishna Moorthy and Alan, Conrad Bovik},
   doi = {10.1109/TIP.2012.2214050},
   number = {12},
   journal = {IEEE Transactions on Image Processing},
   pages = {4695--4708},
   pmid = {22910118},
   title = {No-reference image quality assessment in the spatial domain},
   volume = {21},
   year = {2012}
}

@misc{pique,
  title={Blind image quality evaluation using perception based features},
  author={Venkatanath, Narasimhan and Praneeth, D and Bh, Maruthi Chandrasekhar and Channappayya, Sumohana S and Medasani, Swarup S},
  note ={Paper presented at the 2015 Twenty First National Conference on Communications (NCC), Mumbai, India, 27 February -- 01 March 2015},
  year={2015},
  organization={IEEE},
  doi = {10.1109/NCC.2015.7084843}
}

@misc{Zhang2023,
  title={Blind image quality assessment via vision-language correspondence: A multitask learning perspective},
  author={Zhang, Weixia and Zhai, Guangtao and Wei, Ying and Yang, Xiaokang and Ma, Kede},
  note={Paper presented at the proceedings of the IEEE Conference on Computer Vision and Pattern Recognition, Vancouver, Canada, 18--23 June 2023},
  year={2023},
  doi={10.48550/arXiv.2303.14968}
}

@misc{Wang2023,
  title={Exploring CLIP for assessing the look and feel of images},
  author={Wang, Jianyi and Chan, Kelvin CK and Loy, Chen Change},
  note = {Paper presented at the 37th AAAI conference on artificial intelligence, Washington, DC, USA, 7--14 February 2023},
  year={2023},
  doi={10.1609/aaai.v37i2.25353}
}

@misc{Agnolucci2024,
   author = {Agnolucci, Lorenzo and Galteri, Leonardo and Bertini, Marco},
   title = {{Q}uality-{A}ware {I}mage-{T}ext {A}lignment for {O}pinion-{U}naware {I}mage {Q}uality {A}ssessment},
   note={Preprint at \url{https://arxiv.org/abs/2403.11176}},
   year={2024}
}

@misc{arniqa,
      title={ARNIQA: Learning Distortion Manifold for Image Quality Assessment}, 
      author={Agnolucci, Lorenzo and Galteri, Leonardo and Bertini, Marco and Del Bimbo, Alberto},
      year={2023},
      doi={10.1109/WACV57701.2024.00026},
      note = {Paper presented at the 2024 IEEE/CVF Winter Conference on Applications of Computer Vision (WACV), Waikoloa, Hawaii, 3--7 January 2023}
}

@misc{lariqa,
  title={{L}{A}{R}-{I}{Q}{A}: A {L}ightweight, {A}ccurate, and {R}obust {N}o-{R}eference {I}mage {Q}uality {A}ssessment {M}odel},
  author={Avanaki, Nasim Jamshidi and Ghildyal, Abhijay and Barman, Nabajeet and Zadtootaghaj, Saman},
  note = {Paper presented at the 18th European Conference on Computer Vision ECCV, MiCo, Milano, 29 September -- 04 October 2024},
  year={2024},
  doi = {10.48550/arXiv.2408.17057}
}

@article{wang2004image,
  title={Image quality assessment: from error visibility to structural similarity},
  author={Wang, Zhou and Bovik, Alan C and Sheikh, Hamid R and Simoncelli, Eero P},
  journal={IEEE Transactions on Image Processing},
  volume={13},
  number={4},
  pages={600--612},
  year={2004},
  publisher={IEEE},
  doi = {10.1109/TIP.2003.819861}
}

@misc{zhang2018unreasonable,
  title={The unreasonable effectiveness of deep features as a perceptual metric},
  author={Zhang, Richard and Isola, Phillip and Efros, Alexei A and Shechtman, Eli and Wang, Oliver},
  booktitle={Paper presented at the IEEE Conference on Computer Vision and Pattern Recognition, Salt Lake City, Utah, USA, 18--22 June 2018},
  year={2018},
  doi = {10.48550/arXiv.1801.03924}
}

@article{scholkopf2000new,
  title={New support vector algorithms},
  author={Sch{\"o}lkopf, Bernhard and Smola, Alex J and Williamson, Robert C and Bartlett, Peter L},
  journal={Neural computation},
  volume={12},
  number={5},
  pages={1207--1245},
  year={2000},
  doi = {10.1162/089976600300015565},
  publisher={MIT Press One Rogers Street, Cambridge, MA 02142-1209, USA journals-info~…}
}

@misc{ke2021musiq,
  title={Musiq: Multi-scale image quality transformer},
  author={Ke, Junjie and Wang, Qifei and Wang, Yilin and Milanfar, Peyman and Yang, Feng},
  note={Paper presented at the IEEE Conference on Computer Vision and Pattern Recognition, 19--25 June 2021},
  year={2021},
  doi = {10.48550/arXiv.2108.05997}
}

@misc{ying2020patches,
  title={From patches to pictures (PaQ-2-PiQ): Mapping the perceptual space of picture quality},
  author={Ying, Zhenqiang and Niu, Haoran and Gupta, Praful and Mahajan, Dhruv and Ghadiyaram, Deepti and Bovik, Alan},
  booktitle={Paper presented at the Proceedings of the IEEE Conference on Computer Vision and Pattern Recognition, 14--19 June 2020},
  year={2020},
  doi = {10.48550/arXiv.1912.10088}
}

@article{madhusudana2022image,
  title={Image quality assessment using contrastive learning},
  author={Madhusudana, Pavan C and Birkbeck, Neil and Wang, Yilin and Adsumilli, Balu and Bovik, Alan C},
  journal={IEEE Transactions on Image Processing},
  volume={31},
  pages={4149--4161},
  year={2022},
  publisher={IEEE},
  doi = {10.1109/TIP.2022.3181496}
}

@misc{saha2023re,
  title={Re-iqa: Unsupervised learning for image quality assessment in the wild},
  author={Saha, Avinab and Mishra, Sandeep and Bovik, Alan C},
  booktitle={Paper presented at the IEEE Conference on Computer Vision and Pattern Recognition, Vancouver, Canada, 18--23 June 2023},
  year={2023},
  doi = {10.48550/arXiv.2304.00451}
}

@misc{liu2024kan,
  title={Kan: Kolmogorov-arnold networks},
  author={Liu, Ziming and Wang, Yixuan and Vaidya, Sachin and Ruehle, Fabian and Halverson, James and Solja{\v{c}}i{\'c}, Marin and Hou, Thomas Y and Tegmark, Max},
  note={Paper presented at the Twelfth International Conference on Learning Representations, Vienna, Austria,
7--11 May 2024},
  year={2024},
  doi = {10.48550/arXiv.2404.19756}
}

@misc{fivek,
	author = "Vladimir, Bychkovsky and Sylvain, Paris and Eric, Chan and Fr{\'e}do, Durand",
	title = "Learning Photographic Global Tonal Adjustment with a Database of Input / Output Image Pairs",
	  note = "Paper presented at the Twenty-Fourth IEEE Conference on Computer Vision and Pattern Recognition, Colorado Springs, CO, USA, 20--25 June 2011 ",
	year = "2011",
    doi = {10.1109/CVPR.2011.5995413}
}

@article{ak2022rv,
  title={{R}{V}-{T}{M}{O}: {L}arge-{S}cale {D}ataset for {S}ubjective {Q}uality {A}ssessment of {T}one {M}apped {I}mages},
  author={Ak, Ali and Goswami, Abhishek and Hauser, Wolf and Le Callet, Patrick and Dufaux, Fr{\'e}d{\'e}ric},
  journal={IEEE Transactions on Multimedia},
  volume={25},
  pages={6013--6025},
  year={2022},
  publisher={IEEE},
  doi = {10.1109/TMM.2022.3203211}
}

@article{jiang2025dataset,
  title={{D}ataset and {M}etric for {Q}uality {A}ssessment of {H}{D}{R} {T}one {M}apping: {D}etail {V}isibility, {C}olor {N}aturalness, and {O}verall {Q}uality},
  author={Jiang, Qiuping and Li, Xiwen and Wang, Xinyi and Wang, Zhihua and Zhai, Guangtao},
  journal={IEEE Transactions on Multimedia},
  year={2025},
  month = {1},
  publisher={IEEE},
  doi={10.1109/TMM.2025.3535338}
}

@article{bianco2018use,
  title={On the use of deep learning for blind image quality assessment},
  author={Bianco, Simone and Celona, Luigi and Napoletano, Paolo and Schettini, Raimondo},
  journal={Signal, Image and Video Processing},
  volume={12},
  number={2},
  pages={355--362},
  year={2018},
  publisher={Springer},
  doi = {10.1007/s11760-017-1166-8}
}

@misc{li2022blip,
  title={BLIP: Bootstrapping language-image pre-training for unified vision-language understanding and generation},
  author={Li, Junnan and Li, Dongxu and Xiong, Caiming and Hoi, Steven},
  note ={Paper presented at the Thirty-Ninth International Conference on Machine Learning, Baltimore, MD, USA, 17--23 July 2022},
  pages={12888--12900},
  year={2022},
  doi = {10.48550/arXiv.2201.12086}
}

@article{mantiuk2012comparison,
  title={{C}omparison of {F}our {S}ubjective {M}ethods for {I}mage {Q}uality {A}ssessment},
  author={Mantiuk, Rafa{\l} K and Tomaszewska, Anna and Mantiuk, Rados{\l}aw},
  journal={Computer graphics forum},
  volume={31},
  number={8},
  pages={2478--2491},
  year={2012},
  doi = {10.1111/j.1467-8659.2012.03188.x}
}

@article{bradley1952rank,
  title={Rank analysis of incomplete block designs: I. The method of paired comparisons},
  author={Bradley, Ralph Allan and Terry, Milton E},
  journal={Biometrika},
  volume={39},
  number={3/4},
  pages={324--345},
  year={1952},
  doi = {10.2307/2334029}
}

@article{parducci1965category,
  title={Category judgment: a range-frequency model.},
  author={Parducci, Allen},
  journal={Psychological review},
  volume={72},
  number={6},
  pages={407--418},
  year={1965},
  publisher={American Psychological Association},
  doi={10.1037/h0022602}
}

@article{tversky1974judgment,
  title={{J}udgment under {U}ncertainty: {H}euristics and {B}iases: {B}iases in judgments reveal some heuristics of thinking under uncertainty.},
  author={Tversky, Amos and Kahneman, Daniel},
  journal={Science},
  volume={185},
  number={4157},
  pages={1124--1131},
  year={1974},
  publisher={American association for the advancement of science},
  doi={10.1126/science.185.4157.1124}
}
